\documentclass[conference]{IEEEtran}

\usepackage{times}
\usepackage{balance}

\usepackage[utf8]{inputenc}
\usepackage[english]{babel}
\usepackage[T1]{fontenc}
\usepackage{epsfig}
\usepackage{graphicx}
\usepackage{amssymb,amsmath}
\usepackage{booktabs}
\usepackage{flushend}
\usepackage{fixltx2e}
\usepackage{tikz}
\usepackage{hyperref}

\newcommand{\ra}[1]{\renewcommand{\arraystretch}{#1}}
\newcommand{\specialcell}[2][c]{%
  \begin{tabular}[#1]{@{}c@{}}#2\end{tabular}}

\begin{document}
%
\title{Writer-independent Feature Learning for Offline Signature Verification using Deep Convolutional Neural Networks}


\author{

\IEEEauthorblockN{Luiz G. Hafemann, Robert Sabourin}
\IEEEauthorblockA{Lab. d'imagerie, de vision et d'intelligence artificielle\\
\'Ecole de technologie sup\'erieure\\
Universit\'e du Qu\'ebec, Montreal, Canada\\
lghafemann@livia.etsmtl.ca, robert.sabourin@etsmtl.ca}

\and

\IEEEauthorblockN{Luiz S. Oliveira}
\IEEEauthorblockA{Department of Informatics\\
Federal University of Parana\\
Curitiba, PR, Brazil\\
lesoliveira@inf.ufpr.br}

\and

}

\usetikzlibrary{calc}
\maketitle
\begin{tikzpicture}[remember picture, overlay]
\node at ($(current page.north) + (0,-0.5in)$) {\Large Accepted as a conference paper for IJCNN 2016};
\end{tikzpicture}

\begin{abstract}
Automatic Offline Handwritten Signature Verification has been researched over the last few decades from several perspectives, using insights from graphology, computer vision, signal processing, among others. In spite of the advancements on the field, building classifiers that can separate between genuine signatures and skilled forgeries (forgeries made targeting a particular signature) is still hard. We propose approaching the problem from a feature learning perspective. Our hypothesis is that, in the absence of a good model of the data generation process, it is better to learn the features from data, instead of using hand-crafted features that have no resemblance to the signature generation process. To this end, we use Deep Convolutional Neural Networks to learn features in a writer-independent format, and use this model to obtain a feature representation on another set of users, where we train writer-dependent classifiers. We tested our method in two datasets: GPDS-960 and Brazilian PUC-PR. Our experimental results show that the features learned in a subset of the users are discriminative for the other users, including across different datasets, reaching close to the state-of-the-art in the GPDS dataset, and improving the state-of-the-art in the Brazilian PUC-PR dataset.

\end{abstract}

\IEEEpeerreviewmaketitle

\section{Introduction}
Biometrics technology is used in a wide variety of security applications.
The aim of such systems is to recognize a person based on physiological traits (e.g fingerprint, iris) or behavioral traits (e.g. voice,  handwritten signature) \cite{jain_introduction_2004}. The handwritten signature is a particularly important type of biometric trait, mostly due to its widespread use to verify a person's identity in legal, financial and administrative  areas. One of the reasons for its extensive use is that the process to collect handwritten signatures is non-invasive, and people are familiar with their use in daily life \cite{plamondon_online_2000}.

Research in signature verification is divided between online (dynamic) and offline (static) scenarios. In the online case, the signature is captured using a special input device (such as a tablet), and the dynamic information of the signature process is captured (pen's position, inclination, among others). In this work, we focus on the Offline (static) signature verification problem, where the signature is acquired after the writing process is completed, by scanning the document containing the signature. In this case, the signature is represented as a digital image. 

Most of the research effort in this area has been devoted to obtaining a good feature representation for signatures, that is, designing good feature extractors. To this end, researchers have used insights from graphology, computer vision, signal processing, among other areas \cite{hafemann_offline_2015}. As with several problems in computer vision, it is often hard to design good feature extractors, and the choice of which feature descriptors to use is problem-dependent. Ideally, the features should reflect the process used to generate the data - for instance, neuromotor models of the hand movement. Although this approach has been explored in the context of online signature verification \cite{ferrer_static_2015}, there is not a widely accepted ``best'' way to model the problem, specially for Offline (static) signature verification, where the dynamic information of the signature generation process is not available. 

\begin{figure}
\includegraphics[width=\columnwidth]{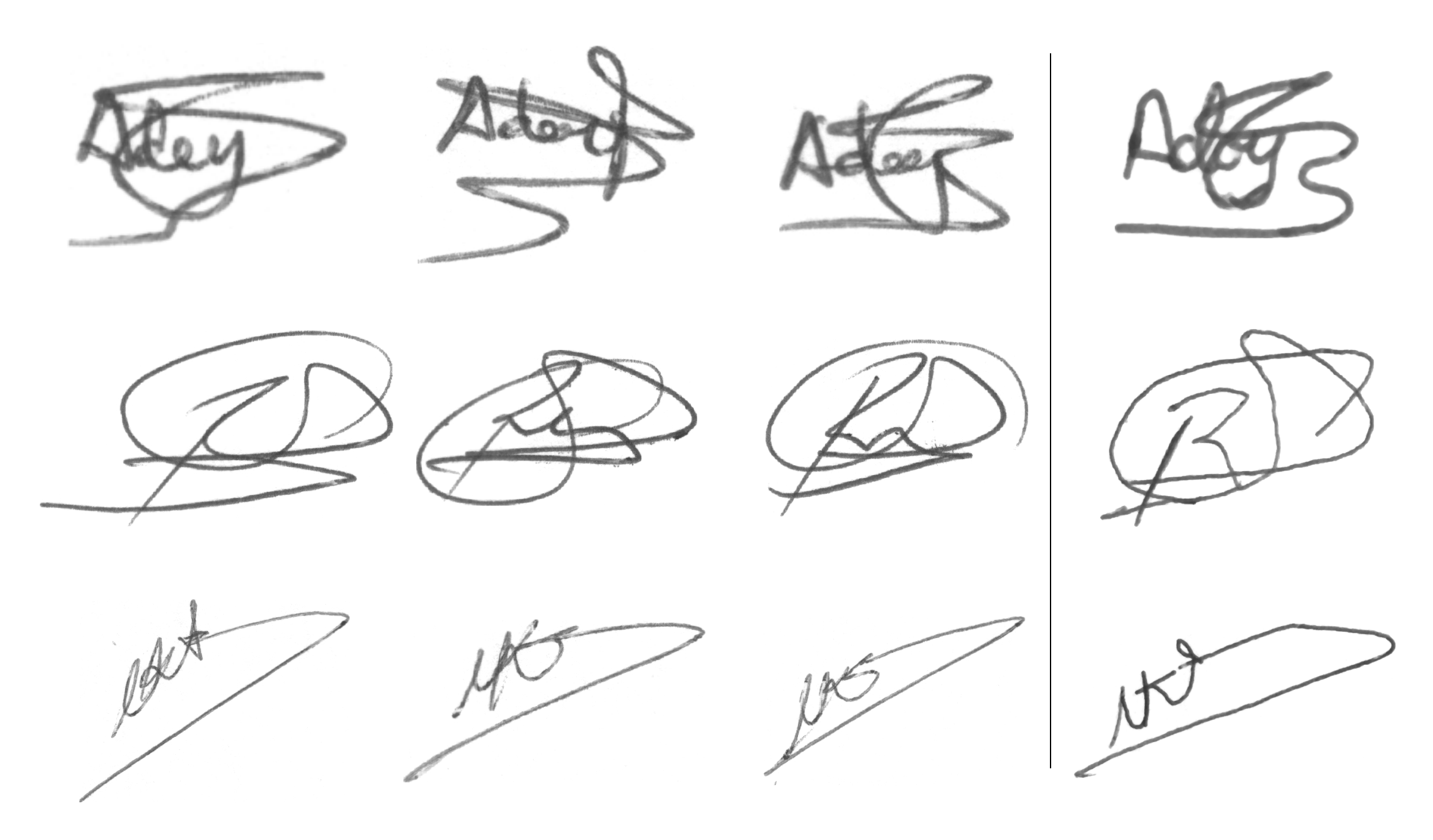}
\caption{Samples from the GPDS-960 dataset. Each row contains three genuine signatures from the same user and a skilled forgery. We notice that each genuine signature is different (showing high intra-class variability), while skilled forgeries resemble the genuine signatures  to a large extent (showing low inter-class variability)}
\label{fig:forgery}
\end{figure}

In spite of the advancements in the field, systems proposed in the literature still struggle to distinguish genuine signatures and skilled forgeries. These are forgeries made by a person with access to a user's signature, that practices imitating it (see Figure \ref{fig:forgery}). Experimental results show somewhat large error rates when testing on public datasets (such as GPDS \cite{vargas_off-line_2007}), even when the number of samples for training is around 10-15 (results are worse with 1-3 samples per user, which is a common scenario in banks and other institutions). 

In this work we propose using feature learning (also called representation learning) for the problem of Offline Signature Verification, in order to obtain better feature representations. Our hypothesis is that, in the absence of a good model of the data generation process, it is better to learn the features from data, rather than using hand-crafted features that have no resemblance to how the signatures are created, which is the case for the best performing  systems proposed in the literature. For example, recent Offline Signature Verification systems are based on texture descriptors, such as Local Binary Patterns \cite{serdouk_off-line_2015}, interest-point-matching such as SURF \cite{pal_off-line_2012}, among others.

We base our research on recent successful applications of purely supervised learning models for computer vision (such as image recognition \cite{krizhevsky_imagenet_2012}). In particular, we use Deep Convolutional Neural Networks (CNN) trained with a supervised criterion, in order to learn good representations for the signature verification problem. This type of architecture is interesting for our problem, since it scales better than fully connected models for larger input sizes, having a smaller number of trainable parameters. This is a desirable property for the problem at hand, since we cannot rescale signature images too much without risking losing the details that enable discriminating between skilled forgeries and genuine signatures. 

The most common formulation of the signature verification problem is called Writer-Dependent classification. In this formulation, one classifier is built for each user in the system. Using a supervised feature learning approach directly in this case is not practical, since the number of samples per user is very small (usually around 1-14 samples). Instead, we propose a two-phase approach: a Writer-Independent feature learning phase followed by Writer-Dependent classification. The feature learning phase uses a surrogate classification task for learning feature representations, where we train a CNN to discriminate between signatures from users not enrolled in the system. We then use this CNN as a feature extractor and train a Writer-dependent classifier for each user. Note that in this formulation, adding a new user to the system requires training only a Writer-Dependent classifier.

We tested this method using two datasets: the GPDS-960 corpus (\cite{vargas_off-line_2007}) and the Brazilian PUC-PR dataset \cite{freitas_bases_2000}. The first is the largest publicly available corpus for offline signature verification, while the second is a smaller dataset that has been used for several studies in the area. 

Our main contributions are the following: We propose a two-stage framework for offline signature verification, where we learn features in a Writer-Independent way, and build Writer-Dependent classifiers. Our results show that we do have enough data in signature datasets to learn relevant features for the task, and the proposed method achieves state-of-the-art performance. We also investigate how the features learned in one dataset transfer to another dataset, and the impact in performance of the number of samples available for WD training.

\section{Related Work}

Feature learning methods have not yet been broadly researched for the task of offline signature verification. Murshed et al. \cite{murshed_binary_1997}, \cite{murshed_cognitive_1997}, used autoencoders (called Identity-Mapping Backpropagation in their work) to perform dimensionality reduction followed by a Fuzzy ARTMAP classifier. This work, however, considered only a single hidden layer, with less units than the input. In contrast, in recent successful applications of autoencoders, multiple layers of representations are learned, often in an over-complete format (more hidden units than visible units), where the idea is not to reduce dimensionality, but ``disentangle'' the factors of variation in the inputs \cite{bengio_learning_2009}. Ribeiro et al. \cite{ribeiro_deep_2011} used unsupervised learning for learning representations, in particular, Restricted Boltzmann Machines (RBMs). In this work, the authors tested with a small subset of users (10 users), and only reported a visual representation of the learned weights, and not the results of using such features to discriminate between genuine signatures and forgeries. Khalajzadeh \cite{khalajzadeh_persian_2012} used Convolutional Neural Networks (CNNs) for Persian signature verification, but did not considered skilled forgeries.

A similar strategy to our work has been used by Sun et al. \cite{sun_deep_2014} for the task of face verification. They trained CNNs on a large dataset of faces and used these networks to extract features on another face dataset. In their work, the verification process consisted in distinguishing between faces from different users. In signature verification, distinguishing between different writers is one of the objectives (when we consider ``Random Forgeries''), but the main challenge is to distinguish between genuine signatures and skilled forgeries. In this work we evaluate the method for both types of forgery.

The framework we propose is also similar to previous work by Eskander et al. \cite{eskander_hybrid_2013}. In this work, a Writer-Independent set is used for feature selection, and a Writer-Dependent set is used for training and evaluation. However, in this work the authors used hand-crafted feature extractors, while in the present work we use the Writer-Independent set for feature learning, instead of feature selection.

\section{Proposed Method}

We propose a two-stage approach, considering a writer-independent feature learning phase, followed by writer-dependent classification. We start by partitioning the dataset into two distinct sets: Development set $\mathcal{D}$ and Exploitation set $\mathcal{E}$. The set $\mathcal{D}$ is used to learn the feature representation for signatures. We consider this as a separate dataset from the enrolled users. The exploitation set $\mathcal{E}$ considers the users enrolled to the system. This set is used to train Writer-Dependent classifiers (using only genuine signatures) and for evaluating the performance of the system.

\begin{figure*}
\centering
\includegraphics[scale=0.95]{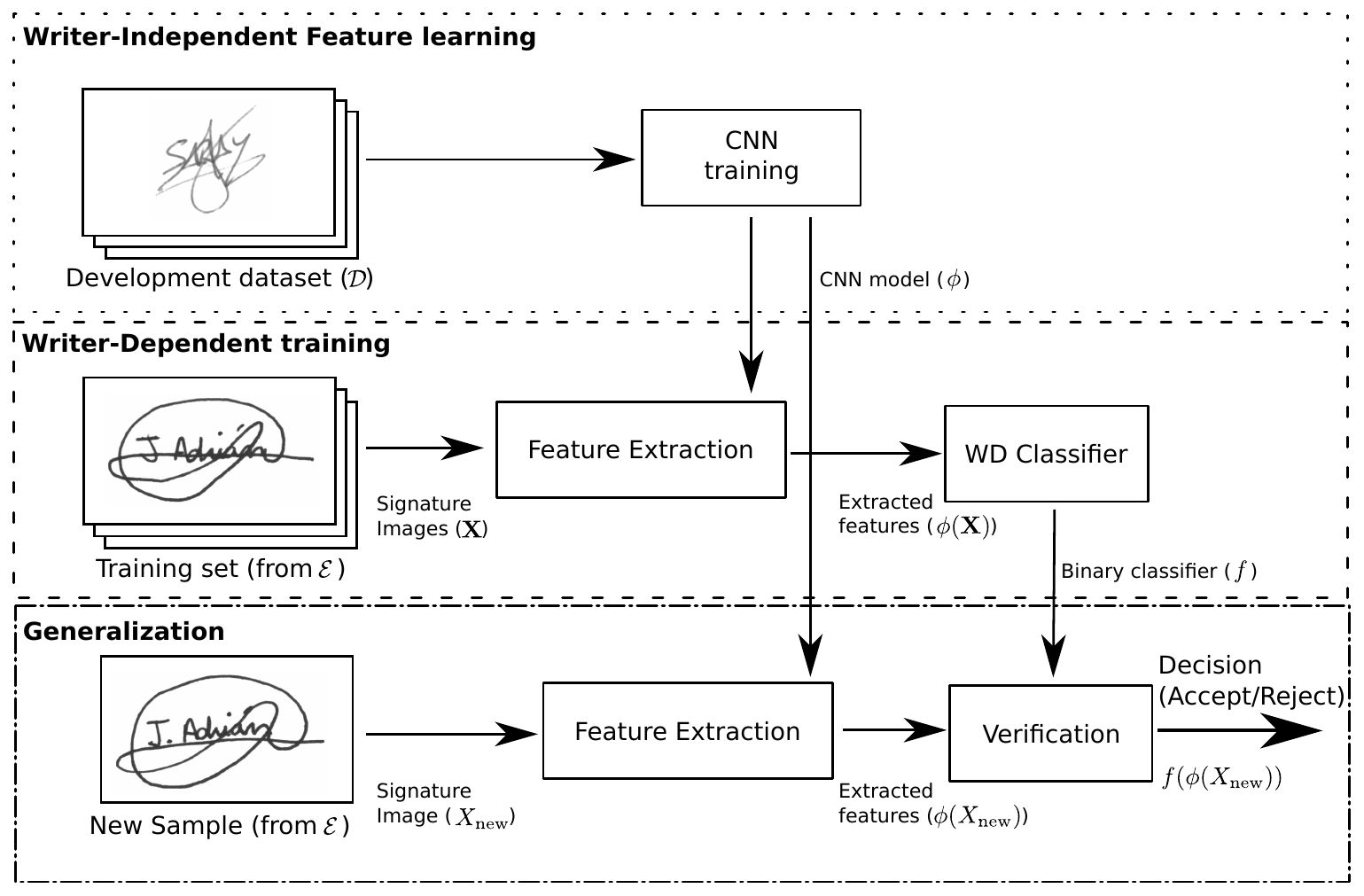}
\caption{The proposed architecture for writer-independent feature learning and writer-dependent classification.}
\label{fig:architecture}
\end{figure*}

The proposed system is illustrated in Figure \ref{fig:architecture}. We first use the set $\mathcal{D}$  to learn the feature representations, by training a Convolutional Neural Network (detailed in the next section). The result is a function $\phi(.)$, learned from data, that projects the input images $X$ to another feature space: $\phi(X) \in \mathbb{R}^m$, where $m$ is the dimensionality of the projected feature space. Our expectation is that the features learned using $\mathcal{D}$ will be useful to separate genuine signatures and forgeries from other users.
After the CNN is trained, we create a training dataset for each user in set $\mathcal{E}$, using a subset of the user's genuine signatures, and random forgeries. We use the CNN as a feature extractor, obtaining a feature vector $\phi(X)$ for each signature $X$ in the user's dataset. This new representation is then  used to train a binary classifier $f$. For a new sample $X_{\text{new}}$, we first use the CNN to ``extract features'' (i.e. obtain the feature vector $\phi(X_{\text{new}})$) and feed the feature vector to the binary classifier, obtaining a final decision $f(\phi(X_{\text{new}}))$. The next sections detail the WI and WD training procedures.

\subsection{Pre-processing}

For all signatures from both datasets ($\mathcal{D}$ and $\mathcal{E}$), we apply the same pre-processing strategy. The signatures from the GPDS dataset have a variable size, ranging from 153 x 258 pixels to 819 x 1137 pixels. Since for training a neural network we need the inputs to have all the same size, we need to normalize the signature images. We evaluated two approaches:

In the simplest approach, we resized the images to a fixed size, using bi-linear interpolation. We perform  rescaling without deformations, that is, when the original image had a different width-to-height ratio, we cropped the excess in the larger dimension.

The second approach consisted in first normalizing the images to the largest image size, by padding the images with white background. In this case, we centered the signatures in a canvas of size 840 x 1360 pixels, aligning the center of mass of the signature to the center of the image, similar to previous approaches in the literature, e.g. \cite{pourshahabi_offline_2009}. We then rescaled the images to the desired input size of the neural network.

With the first approach, less fine-grained information is lost during the rescaling, specially for the users that have small signatures. On the other hand, the width of the pen strokes  becomes inconsistent: for the smaller signatures the pen strokes become much thicker than the pen strokes from the larger signatures. 

Besides resizing the images to a standard size, we also performed the following pre-processing steps:
\begin{itemize}
\item \textbf{Removed the background}: we used OTSU's algorithm \cite{otsu_threshold_1975} to find the optimum threshold between foreground and background pixel intensities. Pixels with intensity larger than the threshold were set to white (intensity 255). The signature pixels (with intensity less than the threshold) remain unchanged in this step.
\item \textbf{Inverted the images}: we inverted the images so that the white background corresponded to pixel intensity 0. That is, each pixel of the image is calculated as: $I_\text{inverted}(i,j) \leftarrow 255 - I(i,j)$.
\item \textbf{Normalized the input}: we normalized the input to the neural network by dividing each pixel by the standard deviation of all pixel intensities (from all images in $\mathcal{D}$). We do not normalize the data to have mean 0 (another common pre-processing step) since we want the background pixels to be zero-valued.
\end{itemize}

\subsection{Writer-Independent feature learning}

For learning the representation for signatures, we used Deep Convolutional Neural Networks. We note that modeling directly the problem of interest is not feasible in practice: our ultimate goal is to separate genuine signatures from skilled forgeries of the users enrolled in the system, but in a realistic scenario we only have genuine signatures provided during an enrollment phase, and do not have forgeries for these users. 
Therefore, we need to consider a surrogate classification objective. In this work, we use a separate set of users (the development set $\mathcal{D}$) to learn the features, by learning a classification task, considering each user in $\mathcal{D}$ as a different class. The objective function is to minimize a cross-entropy classification loss. The expectation is that by learning to distinguish between signatures from different users in this dataset, the network will learn features that are relevant for our problem of interest - separating genuine signatures and forgeries from the exploitation set $\mathcal{E}$. 

We used a CNN architecture similar to the one defined by Krizhevsky et al. \cite{krizhevsky_imagenet_2012} for an image recognition problem. Initial tests showed that the capacity of this network seems to be too large for the problem at hand, particularly considering the fully connected layers (that contain most of the weights in the network). We obtained better results with 2 fully-connected layers after the convolutions, instead of three layers from the original model.
For the purpose of replicating our experiment, we provide a full list of the parameters used in our tests.  Table \ref{table:cnn_architecture} lists the definition of the CNN layers. For convolution and pooling layers, we list the size as $N$x$H$x$W$ where N is the number of filters, H is the height and W is the width of the convolution and pooling windows, respectively. \textbf{Stride} refers to the distance between applications of the convolution (or pooling) operation, and \textbf{pad} refers to add padding (borders) to the input, with value $0$. Local Response normalization is applied according to \cite{krizhevsky_imagenet_2012}, with the parameters listed in the table. For the first two fully-connected layers we use dropout \cite{hinton_improving_2012}, with rate 0.5. We use Rectified Linear Units (ReLUs) as the activation function for all convolutional and fully-connected layers, except the last one. The last layer uses a softmax activation and has $N$ neurons, where $N$ is the number of users in the set $\mathcal{D}$, indicating the probability of the sample belonging to each of the users.%

\begin{table} \centering
\caption{Summary of the CNN layers}
\begin{tabular}{@{}lcc@{}} \toprule
Layer & Size & Other Parameters \\ \midrule

Convolution & 96x11x11 & stride = 4, pad=0 \\
Local Response Norm. & - & \specialcell{$\alpha = 10^{-4}$, $\beta=0.75$,\\ $k=2$, $n = 5$} \\
Pooling & 96x3x3 & stride = 2\\

Convolution & 256x5x5 & stride = 1, pad=2 \\
Local Response Norm. & - & \specialcell{$\alpha = 10^{-4}$, $\beta=0.75$,\\ $k=2$, $n = 5$} \\
Pooling & 256x3x3 & stride = 2\\

Convolution & 384x3x3 & stride = 1, pad=1 \\
Convolution & 256x3x3 & stride = 1, pad=1 \\
Pooling & 256x3x3 & stride = 2 \\
Fully Connected + Dropout & 4096 & $p = 0.5$ \\
Fully Connected + Softmax & N & \\

\bottomrule
\end{tabular}
\label{table:cnn_architecture}
\end{table}

We initialize the weights of the model according to the work of Glorot and Bengio \cite{glorot_understanding_2010}, and the biases to $0$. We trained the model with Nesterov Momentum for $60$ epochs, using momentum rate of 0.9, and mini-batches of size $100$. We started with a learning rate of $0.01$, and divided it by 10 twice (after 20 epochs, and after 40 epochs). We used L2 regularization with a weight decay factor of $0.0005$. These values are consolidated in Table \ref{table:otherparameters}. The networks were trained using the libraries Theano \cite{bergstra_theano:_2010} and Lasagne  \cite{dieleman_lasagne:_2015}, and took around 5h to train on a GPU Tesla C2050.

\begin{table} \centering
\caption{Training Hyperparameters}
\begin{tabular}{@{}lcc@{}} \toprule
Parameter & value \\ \midrule
Initial Learning Rate (LR) & 0.01 \\
Learning Rate schedule & $LR \leftarrow LR * 0.1$ (every 20 epochs) \\
Weight Decay & 0.0005 \\
Momentum & 0.9 \\
Batch size & 100 \\

\bottomrule
\end{tabular}
\label{table:otherparameters}
\end{table}

\subsection{Writer-dependent classification}

After the CNN is trained in the set $\mathcal{D}$, we use it to extract features for the Writer-Dependent training. Similar to previous work in transfer learning  \cite{oquab_learning_2014}, \cite{hafemann_transfer_2015}, we use the representation obtained by performing forward propagation of an input image until the last layer before softmax. In the notation defined above, we consider our feature extractor function $\phi(X)$ to be the representation of the network at the last layer before softmax, after forward propagating the input $X$. As noted in Table \ref{table:cnn_architecture}, this representation has 4096 dimensions ($\phi(X) \in \mathcal{R}^{4096}$). The hypothesis is that the features learned for the set $\mathcal{D}$, during the CNN training, will be relevant for signatures for other users (from the exploitation set).

For training the Writer-Dependent classifiers, no skilled forgeries are used during training or validation, to simulate the scenario for a real application. Following previous work on Writer-Dependent classification, we create a dataset for each user, consisting of genuine signatures and random forgeries (using signatures from other users, from $\mathcal{D}$).

For each user in $\mathcal{E}$, we build a Writer-Dependent training and testing set. The training set is composed of a subset of genuine signatures for the user (as the positive examples), as well as genuine signatures from other users from the development dataset (as the negative examples). The testing set consists of genuine signatures from the user (not used for training), and the skilled forgeries made for the user. With this dataset, we first use the CNN to extract the features for each signature image (that is, compute $\phi(X)$ for each signature $X$). We then train a standard two-class classifier $f$ for each user.

For the WD classification, we test both linear SVMs and SVMs with the RBF kernel \cite{cortes_support-vector_1995}. For the linear SVM, we used the hyperparameter $C = 1$, while for the SVM with RBF kernel we optimize the parameters $C$ and $\gamma$ with a subset of users from the set $\mathcal{D}$ using a grid search. We select the hyperparameters that best classify genuine and skilled forgeries from these users. 

During generalization, for a new signature $X_\text{NEW}$, we first use the CNN to obtain the representation of the signature (i.e. calculate $\phi(X_\text{NEW})$, and then feed this representation to the classifier to obtain a final decision on the sample $f(\phi(X_\text{NEW}))$.

\section{Experimental Protocol}

Feature learning for complex tasks has shown to work better with large datasets. The largest publicly available signature dataset is GPDS-960 \cite{vargas_off-line_2007}, and therefore it is particularly suitable for our proposed method. This dataset contains 24 genuine signatures and 30 forgeries per user, from 881 users, which were captured in a single session \cite{vargas_off-line_2007}. We also tested with a smaller dataset, that also has been extensively used for offline signature verification: the Brazilian PUC-PR dataset \cite{freitas_bases_2000}. This dataset contains signatures from 168 users, and forgeries for the first 60 users.

\begin{figure}
\centering
\includegraphics[scale=0.6]{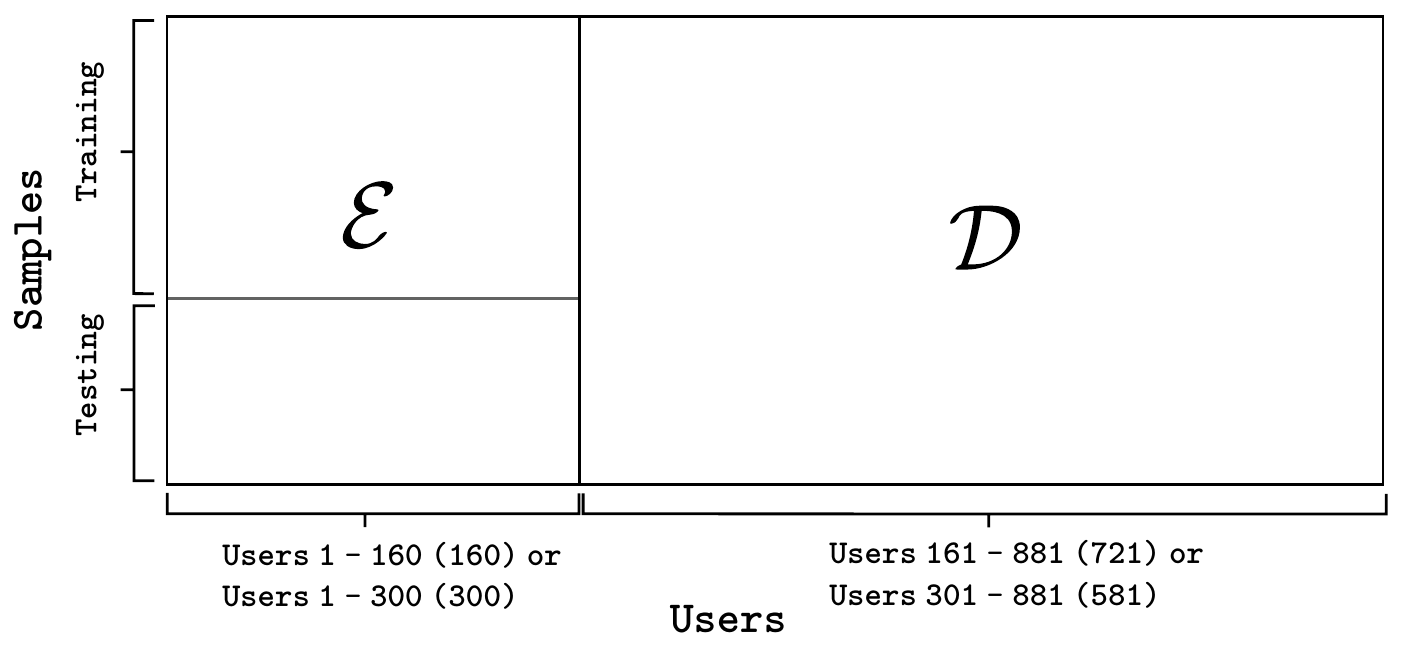}
\caption{The separation of the GPDS-960 dataset in Development set $\mathcal{D}$ and Exploitation set $\mathcal{E}$.}
\label{fig:dataset_separation}
\end{figure}

\begin{table*} \centering
\caption{Training and testing set-up}
\ra{1.3}
\begin{tabular}{@{}ccccc@{}} \toprule

Dataset &\multicolumn{2}{c}{Training Set} & \multicolumn{2}{c}{Testing Set}\\
\cmidrule{2-3}  \cmidrule{4-5} 
&genuine  & forgeries (random) & genuine & forgeries\\ \midrule
Brazilian (PUC-PR) &1,5,10,15,30 samples & 108 x 30 = 3240 samples & 10 samples & 10 random, 10 simple, 10 skilled\\
GPDS-160 &4, 8, 12, 14 samples & 721 x 14 = 10094 samples & 10 samples & 30 skilled\\
GPDS-300 &4, 8, 12, 14 samples & 581 x 14 = 8134 samples & 10 samples &  30 skilled\\
\bottomrule
\end{tabular}
\label{table:dataset}
\end{table*}

The first step is to split the datasets into development set $\mathcal{D}$ and exploitation set $\mathcal{E}$. For GPDS, in order to allow comparison with previous work, we tested with the set $\mathcal{E}$ consisting of the first 160 users, and the first 300 users (which were previously published as GPDS-160 and GPDS-300). Figure \ref{fig:dataset_separation} shows how the dataset is split. The remaining users are used for the writer-independent feature learning phase. For the brazilian set, we consider the first 60 users for the set $\mathcal{E}$, and the remaining 108 users are used as the set $\mathcal{D}$.

After splitting the dataset into sets $\mathcal{D}$ and $\mathcal{E}$, we preprocess the signature images to a standard size of 155 x 220 pixels, considering the two preprocessing options listed in the previous section. This size was chosen to be large enough to keep details from the pen strokes in the signatures, while still small enough to enable training on the GPU. We use the set $\mathcal{D}$ to train a CNN, that learns to classify input signatures to the different users in this set. 

To assess if the learned features generalize to other datasets, we used the CNN trained in the GPDS dataset for extracting features for the brazilian dataset. This experiment serves two purposes: analyze if the learned features generalize to other datasets, and evaluate if we can obtain a better performance on the brazilian set (which is smaller) by leveraging data from a larger dataset (GPDS).

For the Writer-Dependent training, we have slightly different protocols for GPDS and the Brazilian dataset, to correspond to protocols used in other work on these datasets. For GPDS, we selected up to 14 genuine signatures as positive samples (from $\mathcal{E}$), and 14 genuine signatures from each user in the set $\mathcal{D}$ as negative samples. For testing, we selected 10 genuine signatures from the user, ensuring they were not used for training, and all the 30 skilled forgeries. For the Brazilian dataset, we selected up to 30 genuine samples as positive samples (from $\mathcal{E}$), and 30 genuine samples from the users in set $\mathcal{D}$ as negative samples. For testing, we selected 10 genuine signatures from the user, 10 signatures from other users in $\mathcal{E}$ (i.e. not used for training) as random forgeries, and all 10 simple forgeries and 10 skilled forgeries available for each user. 

To evaluate the impact of different number of sample signatures per user, we trained the WD classifiers using a variable number of signatures from the enrolled users. This set-up is summarized in table \ref{table:dataset}.

For optimizing the hyperparameters for the SVM training (for the WD classifiers), we performed a grid search on the parameters $C$ and $\gamma$. We used 10 users from $\mathcal{D}$, building WD classifiers with the same protocol as above. We selected the hyperparameters that performed best in separating genuine signatures and skilled forgeries for these 10 users, by measuring the classification error of each classifier. Before training the SVM models, we rescale the inputs to have a unit standard deviation (in each dimension). This slightly improved performance and significantly decreased the SVM training time. Similar to \cite{eskander_hybrid_2013}, in order to have a balanced dataset for training, we duplicated the genuine examples in the training set to match the same number of random forgeries (equivalent to having different $C$ for the positive and negative classes).

In this work we conducted experiments with two datasets, and authors from different studies have reported different metrics. For GPDS, some authors report two metrics: False Rejection Rate (FRR) and False Acceptance Rate for skilled forgeries (FAR\textsubscript{skilled}). The first metric is the fraction of genuine signatures that were classified as forgery, while the second is the fraction of skilled forgeries that were classified as genuine signatures. Other authors report simply the Equal Error Rate, which is the point in a ROC curve where FAR and FRR are equal. For the results on GPDS, we report these three metrics, and also the mean of the Area Under the Curve (AUC) - that is, we build a ROC curve for each user, and report the average of the AUC. For calculating the EER, we considered the ROC curves created for each user (thresholds specific for each user).

For the Brazilian PUC-PR dataset, authors commonly report FRR and FAR for three types of forgeries: Random, Simple and Skilled. Authors also report an average error rate (AER) which is the average of the four types of error (FRR, FAR\textsubscript{random}, FAR\textsubscript{simple}, FAR\textsubscript{skilled}). To allow comparison with the results on GPDS, we also report metrics considering only FRR and FAR\textsubscript{skilled}: AER\textsubscript{genuine + skilled}, EER\textsubscript{genuine + skilled} and Mean AUC\textsubscript{genuine + skilled}.

\section{Results and Discussion}

We first report the results of the search for the best hyperparameters for the SVM with RBF kernel used for Writer-Dependent classification. After training classifiers for 10 users in the development set, we noticed that the best hyperparameters were the same for most users (8/10 users): $\gamma = 2^{-12}, C = 1$. For the other two users, this was the second best configuration for the parameters. Therefore we used these hyperparameters for the subsequent experiments.

Table \ref{table:gpds_results} presents the results of our experiments with the GPDS dataset. The column ``Features'' list the method we used to extract features - in our work this column lists the CNN trained in the set $\mathcal{D}$. We considered both alternatives defined in the Pre-processing section - simply resizing the signatures images (CNN\_GPDS), and first normalizing the signatures in a canvas with a standard size, before resizing them (CNN\_GPDS\textsubscript{norm}). We notice that this normalization was essential to obtain good classification results on this dataset, with a boost in performance from 14.64\% of EER to 10.70\% in the GPDS-160 dataset. We also noticed the best results were achieved with the SVM trained with an RBF kernel. Lastly, we noted a drop in performance between the experiments with GPDS-160 and GPDS-300. This can be partially explained by the fact that we use more data on the set $\mathcal{D}$ for GPDS-160.

\begin{table}
\centering
\caption{Classification errors on GPDS (\%) and mean AUC}
\label{table:gpds_results}
\resizebox{\columnwidth}{!}{%
\begin{tabular}{lllrrrr}
\toprule
Dataset & Features & Classifier &   FRR &  FAR &  EER &  Mean AUC \\
\midrule
 GPDS-160 &  CNN\_GPDS &  SVM (Linear) & 26.62 &         9.65 &               14.35 &    0.9153 \\
 GPDS-160 &  CNN\_GPDS &     SVM (RBF) & 37.25 &         3.66 &               14.64 &    0.9097 \\
 GPDS-160 &             CNN\_GPDS\textsubscript{norm} &  SVM (Linear) & 11.12 &        16.77 &               11.32 &    0.9381 \\
 \textbf{GPDS-160} &             \textbf{CNN\_GPDS\textsubscript{norm}} &     \textbf{SVM (RBF)} & \textbf{19.81} &         \textbf{5.99} &               \textbf{10.70} &    \textbf{0.9459} \\
 GPDS-300 &  CNN\_GPDS &  SVM (Linear) & 25.43 &        12.80 &               16.40 &    0.8968 \\
 GPDS-300 &  CNN\_GPDS &     SVM (RBF) & 36.27 &         5.00 &               16.22 &    0.9014 \\
 GPDS-300 &             CNN\_GPDS\textsubscript{norm} &  SVM (Linear) & 11.93 &        25.58 &               16.07 &    0.8957 \\
 \textbf{GPDS-300} &             \textbf{CNN\_GPDS\textsubscript{norm}} &     \textbf{SVM (RBF)} & \textbf{20.60} &         \textbf{9.08} &               \textbf{12.83} &    \textbf{0.9257} \\

\bottomrule
\end{tabular}}
\end{table}

Table \ref{table:results_brazilian} shows the results of our tests with the Brazilian PUC-PR dataset. We noticed the same characteristics as with the GPDS test, with improved results with the non-linear RBF kernel for the classifier. In this dataset we tested with both a CNN trained on the brazilian dataset, as well as the CNN trained above for the GPDS dataset. The results were similar, suggesting that the features learned in one dataset generalize well to other datasets. On the other hand, we expected the performance with the CNN trained on GPDS to be better, since the development set for the Brazilian dataset is much smaller (108 users in the Brazilian dataset vs. 721 users for GPDS-160), and therefore there is much more data on GPDS to learn a good feature representation.

\begin{table*}
\centering
\caption{Classification errors on the Brazilian PUC-PR dataset (\%) and mean AUC}
\label{table:results_brazilian}

\begin{tabular}{llrrrrrr}
\toprule
Features & Classifier &  FRR &  FAR\textsubscript{random} &  FAR\textsubscript{simple} &  FAR\textsubscript{skilled} &  EER\textsubscript{genuine + skilled} &  Mean AUC\textsubscript{genuine + skilled} \\
\midrule
       CNN\_Brazilian &  SVM (Linear) & 1.00 &        0.00 &        1.67 &        27.17 &                7.33 &    0.9668 \\
       \textbf{CNN\_Brazilian} &     \textbf{SVM (RBF)} & \textbf{2.83} &        \textbf{0.17} &        \textbf{0.17} &        \textbf{14.17} &                \textbf{4.17} &    \textbf{0.9837}
 \\
 CNN\_GPDS &  SVM (Linear) & 1.83 &        0.00 &        1.33 &        27.83 &               11.50 &    0.9413 \\
 CNN\_GPDS &     SVM (RBF) & 6.50 &        0.17 &        1.17 &        15.17 &                8.50 &    0.9601 \\
            CNN\_GPDS\textsubscript{norm} &  SVM (Linear) & 0.17 &        0.00 &        1.67 &        29.00 &                6.67 &    0.9653 \\
            \textbf{CNN\_GPDS\textsubscript{norm}} &     \textbf{SVM (RBF)} & \textbf{2.17} &        \textbf{0.17} &        \textbf{0.50} &        \textbf{13.00} &                \textbf{4.17} &    \textbf{0.9800} \\

\bottomrule
\end{tabular}

\end{table*}

\begin{figure}
\centering
\includegraphics[width=\columnwidth]{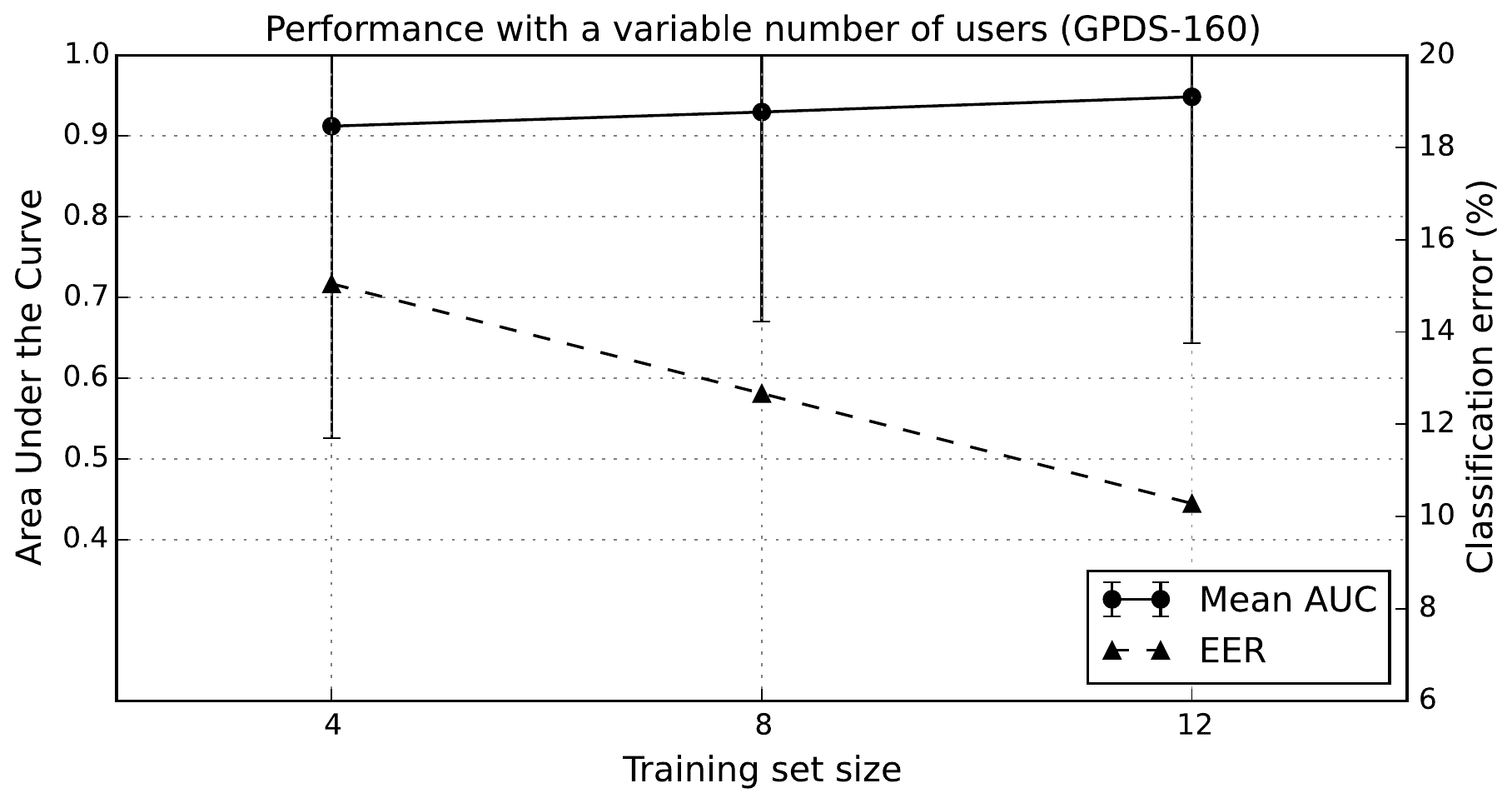}
\caption{Performance on the GPDS-160 dataset varying the number of samples per user for WD training. The error bars show the smallest and largest AUC of users in the exploitation dataset.}
\label{fig:auc_gpds}
\end{figure}

\begin{figure}
\centering
\includegraphics[width=\columnwidth]{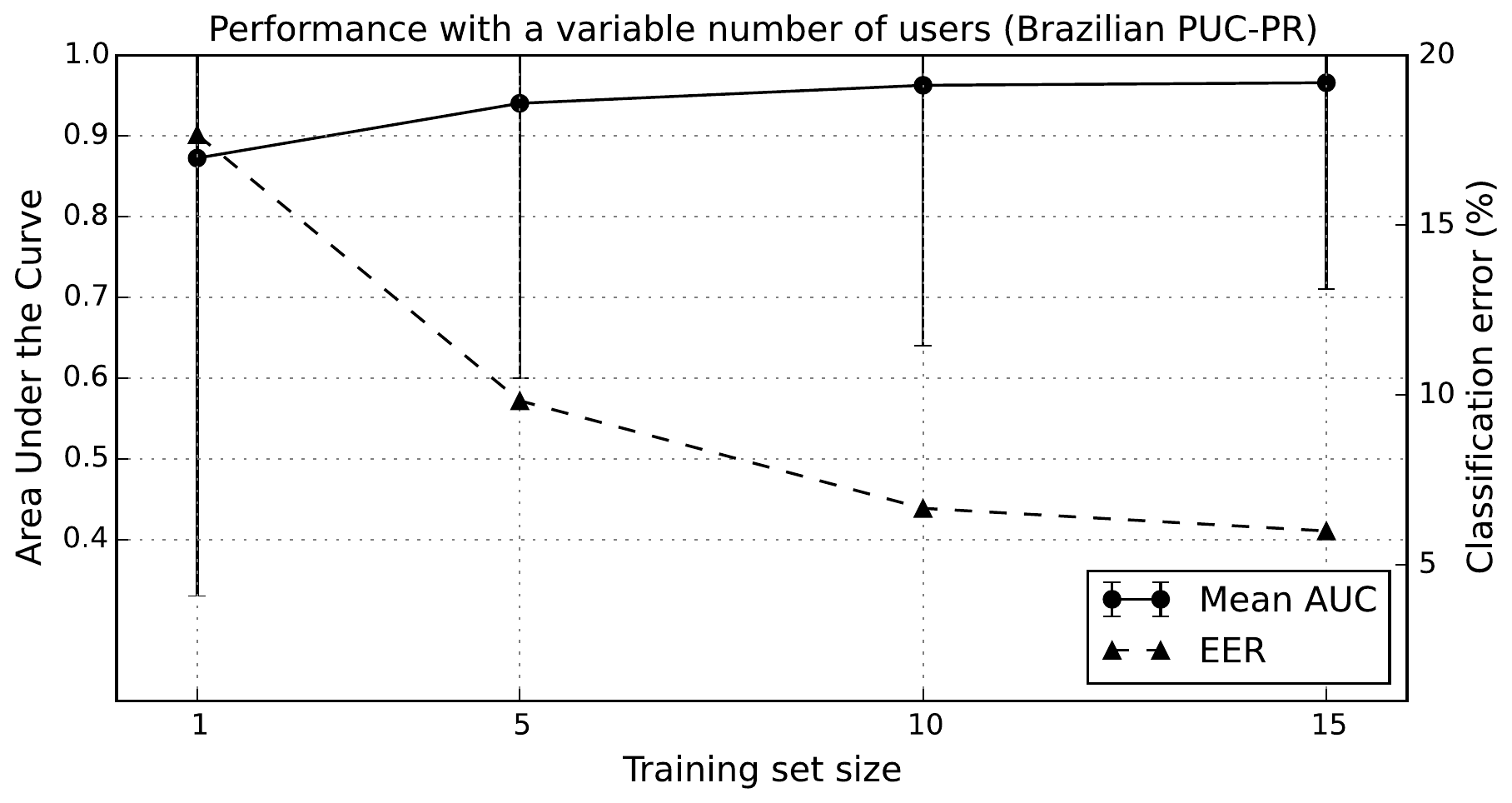}
\caption{Performance on the Brazilian PUC-PR dataset varying the number of samples per user for WD training. The error bars show the smallest and largest AUC of users in the exploitation dataset.}
\label{fig:auc_brazilian}
\end{figure}

We evaluated the performance of the system considering different number of samples per user in the exploitation set. For these tests, we used the configuration that performed best on the tests above: using the normalized GPDS development set to learn the features, and using an SVM with RBF kernel for training the WD classifiers. Figures \ref{fig:auc_gpds} and \ref{fig:auc_brazilian}   present the evolution of the AUC and the Equal Error Rate for the GPDS and Brazilian datasets. We notice that even with a small number of samples the performance is reasonable, achieving 15.05\% EER with 4 signatures in the GPDS dataset, and 9.83\% EER with 5 signatures on the Brazilian dataset. However, we notice that in the extreme case, when a single signature is available, the performance of the entire system is much worse (around 17\% EER), and some users have very poor performance (for one user, AUC is below 0.5).

\begin{table*}
\centering
\caption{Comparison with the state-of-the-art on the Brazilian PUC-PR dataset (errors in \%)}
\label{table:soa_brazilian}
\resizebox{\textwidth}{!}{%

\begin{tabular}{lllrrrrrrr}
\toprule
Reference & Features & Classifier &  FRR &  FAR\_random &  FAR\_simple &  FAR\_skilled &  AER & AER\textsubscript{genuine + skilled} &  EER\textsubscript{genuine + skilled}  \\
\midrule
Bertolini et al. \cite{bertolini_reducing_2010} & Graphometric & SVM (RBF)&10.16&3.16&2.8&6.48&5.65&8.32 & - \\
Batista et al. \cite{batista_dynamic_2012} & Pixel density & HMM + SVM&7.5&0.33&0.5&13.5&5.46&10.5 & -\\
Rivard et al. \cite{rivard_multi-feature_2013} &ESC + DPDF & Adaboost &11&0&0.19&11.15&5.59&11.08& -\\
Eskander et al. \cite{eskander_hybrid_2013}&ESC + DPDF &Adaboost&7.83&0.02&0.17&13.5&5.38&10.67& -\\

\midrule

\textbf{Present Work} & \textbf{CNN\_GPDS\textsubscript{norm}} & \textbf{SVM (RBF)} & \textbf{2.17} &        \textbf{0.17} &        \textbf{0.50} &        \textbf{13.00} & \textbf{3.96} & \textbf{7.59} &             \textbf{4.17} \\
\bottomrule
\end{tabular}
}
\end{table*}

\begin{table}
\centering
\caption{Comparison with state-of-the art on GPDS-160 (errors in \%)}
\label{table:soa_gpds}
\resizebox{\columnwidth}{!}{%
\begin{tabular}{lllrrr}
\toprule
Reference & Features & Classifier &   FRR &  FAR &  EER\\
\midrule

Hu and Chen \cite{hu_offline_2013} & LBP, GLCM, HOG &Adaboost&-&-&7.66\\
Yilmaz \cite{yilmaz_offline_2015} & LBP & SVM (RBF) &-&-&9.64\\
Yilmaz \cite{yilmaz_offline_2015} & LBP, HOG  &Ensemble of SVMs&-&-&6.97\\
Guerbai et al. \cite{guerbai_effective_2015} & 
Curvelet transform & OC-SVM &12.5&19.4&-\\
\midrule

Present work & CNN\_GPDS\textsubscript{norm} & SVM (RBF) & 19.81 &         5.99 &                10.70\\
\bottomrule
\end{tabular}
}
\end{table}

We compare our results with the state-of-the art in tables \ref{table:soa_brazilian} and \ref{table:soa_gpds}. For GPDS, the method achieves state-of-the-art performance in terms of Equal Error Rate, when comparing with systems that used a single feature extractor. However, the performance is worse compared to systems where multiple feature extractors / classifiers are used. Future work can be done in analyzing if the features learned from data are complementary to hand-crafted features.

For the Brazilian PUC-PR dataset, authors use other metrics to compare - the False Acceptance rates for different types of forgery and the Average Error Rate among all types of error. Besides using these metrics, we also compare with an average error rate considering only genuine signatures and skilled forgeries, which is more comparable to the results on GPDS. In this dataset, the proposed method achieves state-of-the-art performance. The large gap between AER\textsubscript{genuine + skilled} and EER\textsubscript{genuine + skilled} also shows that optimization of user-specific decision thresholds is necessary to obtain a good system:  in the present work the decision thresholds were kept as default (scores larger than 0 were considered forgeries). We notice that, for GPDS, this default threshold achieved a large FRR, with low FAR, while for the Brazilian dataset we obtained the opposite. This suggests that a global threshold is not sufficient, and user-specific thresholds should be considered. Better user-specific thresholds will be explored in future work.  

It is worth noting that in the present work we trained the WD classifiers with a combination of genuine signatures and random forgeries. This considers a hypothesis that separating random forgeries from genuine signatures will also make the classifier separate genuine signatures from skilled forgeries. This is a weak hypothesis, as we expect the skilled forgeries to have much more resemblance to the genuine signatures, where random forgeries should be quite different. However, given that we only have genuine signatures available for training, this is a reasonable option, and has been used extensively in the literature for Writer-Dependent classification. An alternative is to use one-class classification to model only the distribution of the genuine signatures (e.g. \cite{guerbai_effective_2015}), which can be explored as future work. 

We would like to point out that, although the EER metric (Equal Error Rate) is useful to have a single number to compare different systems, it relies on implicitly selecting the decision thresholds using information from the test set. Therefore, it considers the error rate that can be achieved with the optimal decision threshold for each user. In a real application, the decision thresholds can only be defined using data from the enrolled users (i.e. using only genuine signature from the training/validation set), or in a writer-independent way (a single global threshold). Therefore, besides reporting EER, we consider beneficial to also report FAR and FRR, stating the procedure used to select the thresholds.

Lastly, we would like to point out that the WD training datasets are significantly imbalanced. We have only a few positive samples(1-30), and a large amount of random forgeries (up to 10 thousand for GPDS-160). Methods betters suited for such scenario can also be explored in future work to improve the performance of the system.

\section{Conclusion}

We presented a two-stage framework for offline signature verification, based on writer-independent feature learning and writer-dependent classification. This method do not rely on hand-crafted features, but instead learn them from data in an writer-independent format. Experiments conducted on GPDS and the Brazilian PUC-PR datasets demonstrate that this method is promising, achieving performance close to the state-of-the-art for GPDS and surpassing the state-of-the-art performance in the Brazilian PUC-PR dataset. We have shown that the features seem to generalize well, by learning the features in the GPDS dataset and achieving good results on the Brazilian PUC-PR dataset. Results with small number of samples per user also demonstrated that this method can be effective even with few samples per user (4-5 samples).

Lastly, we note that although these methods achieve low Equal Error Rates, the actual False Rejection and False Acceptance rates are very imbalanced, and not stable across multiple users and datasets. This highlights the importance of a good method for defining user-specific thresholds, which we intend to explore in future work.

\section*{Acknowledgment}

This research has been supported by the CNPq grant \#206318/2014-6.

\bibliographystyle{IEEEtran}
\bibliography{biblio}

\end{document}